\documentclass[pdftex,twocolumn,10pt,letterpaper,hidelinks]{article}
\usepackage{graphicx, times}
\usepackage{lipsum}
\usepackage{indentfirst}
\usepackage{enumitem}
\usepackage{titlesec}
\usepackage{amsmath}
\usepackage{mathtools}
\usepackage{caption}
\usepackage{subcaption}
\usepackage{textcomp}
\usepackage{tabularx}
\usepackage{bbold}
\usepackage{xcolor}
\usepackage[normalem]{ulem}
\usepackage[T1]{fontenc}
\usepackage{setspace}
\usepackage[linesnumbered, noend, ruled, vlined]{algorithm2e}
\usepackage{hyperref}

\usepackage[available,functional,reproduced]{usenixbadges}

\makeatletter
\newcommand{\nosemic}{\renewcommand{\@endalgocfline}{\relax}}
\newcommand{\dosemic}{\renewcommand{\@endalgocfline}{\algocf@endline}}
\let\oldnl\nl
\newcommand{\nonl}{\renewcommand{\nl}{\let\nl\oldnl}}
\makeatother

\newcolumntype{S}{>{\small}l}
\newcommand{\newparagraph}[1]{{\vspace{2pt}\noindent\textbf{#1}}}

\SetAlFnt{\small}

\titlespacing*{\section}{0pt}{.8\baselineskip}{.8\baselineskip}
\titlespacing*{\subsection}{0pt}{.5\baselineskip}{.3\baselineskip}

\newcommand\correspondingauthor{\thanks{Corresponding author. mohoney2@wisc.edu}}

\begin{document}

\pagestyle{empty}

\title{Marius: Learning Massive Graph Embeddings on a Single Machine}
\author{{Jason Mohoney\correspondingauthor, Roger Waleffe, Henry Xu\thanks{Currently at Maryland, work done while at UW-Madison.}, Theodoros Rekatsinas, Shivaram Venkataraman}\\
{University of Wisconsin-Madison}}
\date{}

\interfootnotelinepenalty=10000

\maketitle

\thispagestyle{empty}

\begin{abstract}
We propose a new framework for computing the embeddings of large-scale graphs on a single machine. A graph embedding is a fixed length vector representation for each node (and/or edge-type) in a graph and has emerged as the de-facto approach to apply modern machine learning on graphs. We identify that current systems for learning the embeddings of large-scale graphs are bottlenecked by data movement, which results in poor resource utilization and inefficient training. These limitations require state-of-the-art systems to distribute training across multiple machines. We propose Marius, a system for efficient training of graph embeddings that leverages partition caching and buffer-aware data orderings to minimize disk access and interleaves data movement with computation to maximize utilization. We compare Marius against two state-of-the-art industrial systems on a diverse array of benchmarks. We demonstrate that Marius achieves the same level of accuracy but is up to one order of magnitude faster. We also show that Marius can scale training to datasets an order of magnitude beyond a single machine's GPU and CPU memory capacity, enabling training of configurations with more than a billion edges and 550 GB of total parameters on a single machine with 16 GB of GPU memory and 64 GB of CPU memory.  Marius is open-sourced at \url{www.marius-project.org}.
\end{abstract}

\section{Introduction} \label{introduction}

Graphs are used to represent the relationships between entities in a wide array of domains, ranging from social media and knowledge bases~\cite{zafarani2014social, fairchild1988graphic} to protein interactions~\cite{brohee2006evaluation}. Moreover, complex graph analysis has been gaining attention in neural network-based machine learning with applications in clustering~\cite{schaeffer2007graph}, link prediction~\cite{zhang2018link, taskar2003link}, and recommendation systems~\cite{ying2018graph}. However, to apply modern machine learning on graphs one needs to convert discrete graph representations (e.g., traditional edge-list or adjacency matrix) to continuous vector representations~\cite{node2vec}. To this end, learnable {\em graph embedding} methods~\cite{goyal2018graph, cai2018comprehensive, wang2017knowledge} are used to assign each node (and/or edge) in a graph to a specific continuous vector representation such that the structural properties of the graph (e.g., the existence of an edge between two nodes or their proximity due to a short path) can be approximated using these vectors. 
In general, graph embedding models aim to capture the global structure of a graph and are complementary to graph neural networks (GNNs)~\cite{nrl-tutorial}. Graph embedding models are primarily used in {\em link prediction} tasks and can also be used to obtain vector representations that form the input to GNNs.

\begin{figure}[t]
    \centering
    \includegraphics[width=0.48\textwidth]{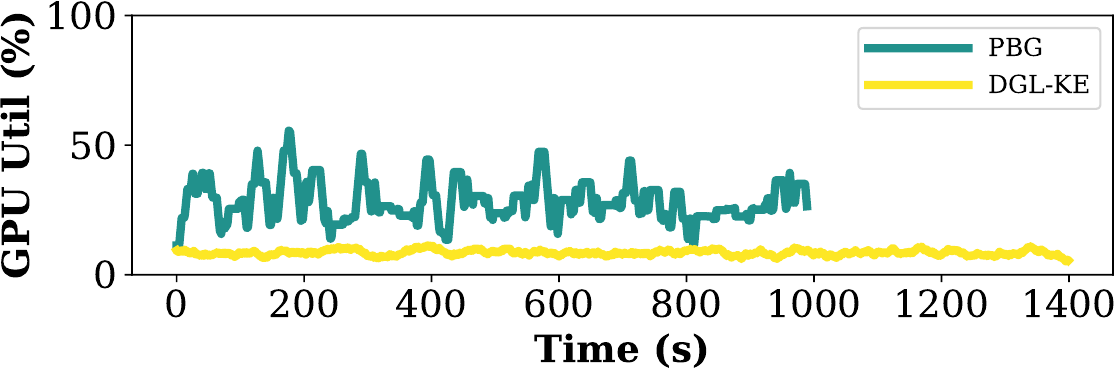}
    \caption{The GPU utilization of DGL-KE and PBG for one training epoch of ComplEx embeddings on the Freebase86m knowledge graph.}
    \label{fig:utilization}
    \vspace{-.1in}
\end{figure}

However, learning a graph embedding model is a resource intensive process. First, training of graph embedding models can be compute intensive: many graph embedding models assign a high-dimensional continuous vector to each node in a graph~\cite{transE, distmult, complex}. For example, it is common to assign a 400-dimensional continuous vector to each node~\cite{pytorchbiggraph, zheng2020dglke}. Consequently, the computational capabilities of GPUs and optimization methods such as mini-batch Stochastic Gradient Descent (SGD) are needed to accelerate training. Second, graph embedding models are memory intensive: the model from our previous example needs 1600 bytes of storage per node and requires 80 GB (the largest GPU memory) for a modest 50 million node graph. Thus, it is necessary to store the learnable parameters in off-GPU memory. Third, the training of graph embedding models requires optimizing over loss functions that consider the edges of the graph as training examples (e.g., the loss can enforce that the cosine similarity between the vector representations of two connected nodes is close to one, see Section~\ref{sec:background}) making training IO-bound for models that do not fit in GPU memory. This limitation arises due to irregular data accesses imposed by the graph structure. As a result, training of large graph embedding models is a non-trivial challenge.

Due to the aforementioned factors, scaling graph embedding training to instances that do not fit in GPU memory introduces costly data movement overheads that can result in poor resource utilization and slow training. In fact, current state-of-the-art systems, including DGL-KE~\cite{zheng2020dglke} from Amazon, and Pytorch BigGraph (PBG)~\cite{pytorchbiggraph} from Facebook, exhibit poor GPU utilization due to these overheads. Figure~\ref{fig:utilization} shows the GPU utilization during a training epoch when using a single GPU for DGL-KE and PBG. As shown, DGL-KE only utilizes 10\% of the GPU, and average utilization for PBG is less than 30\%, dropping to zero during data movement.

GPU under-utilization can be attributed to how these systems handle data movement: To support out-of-GPU-memory training, DGL-KE stores parameters in CPU memory and uses synchronous GPU-based training over minibatches. However, the core computation during graph embedding training corresponds to dot-product operations between vectors (see Section~\ref{preliminaries}), and thus, data transfers dominate the end-to-end run time. Moreover, DGL-KE is fundamentally limited by CPU memory capacity. To address this last limitation, PBG uses a different approach for scaling to large graphs. PBG partitions the embedding parameters into disjoint, node-based partitions and stores them on disk where they can be accessed sequentially. Partitions are then loaded from storage and sent to the GPU where training proceeds synchronously. Doing so avoids copying data from the CPU memory for every batch, but results in GPU underutilization when partitions are swapped.

This problem is exacerbated if the storage device has low throughput. Thus, to scale to large instances both systems opt for distributed training over multiple compute nodes, making training resource hungry. However, the problems these systems face are not insurmountable and can be mitigated. \emph{We show that one can train embeddings on billion-edge graphs using just a single machine.}

We introduce a new pipelined training architecture that can interleave data access, transfer, and computation to achieve high utilization. In contrast to prior systems, our architecture results in high GPU utilization throughout training: for the same workload shown in Figure \ref{fig:utilization}, our approach can achieve an average $\sim70\%$ GPU utilization while achieving the same accuracy (see Section~\ref{sec:experiments}).

To achieve this utilization, our architecture introduces asynchronous training of nodes with \textit{bounded staleness}. We combine this with synchronous training for edge embeddings to handle graphs that may contain edges of different types, for example knowledge graphs where an edge may capture different relationships between nodes. Specifically, we consider learning a separate vector representation for each edge-type. For clarity, we refer to edge-type embeddings as \emph{relation embeddings}. 
This is because updates to the embedding vectors for nodes are sparse and therefore well suited for asynchronous training. However due to the small number of edge-types in real-world graphs ($10,000s$), updates to relation embedding parameters are dense and require synchronous updates for convergence. We design the pipeline to maintain and update node embedding parameters in CPU memory asynchronously, allowing for staleness, while keeping and updating relation embeddings in GPU memory synchronously. Using this architecture, we can train graph embeddings for a billion-edge Twitter graph \emph{one order of magnitude faster} than state-of-the-art industrial systems for the same level of accuracy: Using a single GPU, our system requires 3.5 hours to learn a graph embedding model over the Twitter graph. For the same setting, DGL-KE requires 35 hours.

To scale training beyond CPU memory, 
unlike prior out-of-memory graph processing systems~\cite{10.5555/2387880.2387884}, we need to iterate over edges while computing on data associated with both endpoints.
We propose partitioning the graph and storing embedding parameters on disk. We then design an in-memory \emph{partition buffer} that can hide and reduce IO from swapping of partitions. 
Partitions are swapped from disk into the partition buffer in CPU memory and then used by the training pipeline. Our partition buffer supports pre-fetching and async writes of partitions to hide waiting for IO, resulting in a reduction of training time by up to 2$\times$. Further, we observe that the order in which edge partitions are traversed can impact the number of IOs. Thus, we introduce a \textit{buffer-aware} ordering that uses knowledge of the buffer size and what resides in it to minimize the number of IOs. We show that this 
ordering achieves IO close to the lower bound and provides benefits when compared to locality-based orderings such as Hilbert ordering~\cite{hilbert1891ueber}.

In summary, the key technical contributions of our work are: 1) to show that existing state-of-the-art graph embedding systems are hindered by IO inefficiencies when moving data from disk and from CPU to GPU, 2) to introduce the \emph{Buffer-aware Edge Traversal Algorithm (BETA)}, an algorithm to generate an IO minimizing data ordering for graph learning, 3) to combine the \textit{BETA} ordering with a partition buffer and async IO via pipelining to introduce the first graph learning system that utilizes the full memory hierarchy (Disk-CPU-GPU).

Our design is implemented in Marius, a graph embedding engine that can train billion-edge graphs on a single machine. Using one AWS P3.2xLarge instance, we demonstrate that Marius improves utilization of computational resources and reduces training time by up to an order of magnitude in comparison to existing systems. Marius is 10$\times$ faster than DGL-KE on the Twitter graph with 1.46 billion edges, reducing training times from 35 hours to 3.5 hours. Marius is 1.5$\times$ faster than PBG on the same dataset. On Freebase86m with 86 million nodes and 338 million edges, Marius trains embeddings 3.7$\times$ faster than PBG, reducing training times from 7.5 hours to 2 hours. We also show that Marius can scale to configurations where the parameter size exceeds GPU and CPU memory by an order of magnitude, training a configuration with 550 GB of total parameters, 35$\times$ and 9$\times$ larger than GPU and CPU memory respectively. Finally, we show that despite using a single-GPU on a single-machine, Marius achieves comparable runtime with the multi-GPU configurations of PBG and DGL-KE, thus, providing a cost reduction on cloud resources between 2.9$\times$ and 7.5$\times$ depending on the configuration.

\section{Preliminaries}\label{preliminaries}
We first discuss necessary background on graph embeddings and related systems. Then, we review challenges related to optimizing data movement for training large scale graph embedding models. These are the challenges that this work addresses.
\subsection{Background and Related Work}
\label{sec:background}
\begin{figure}[t]
    \centering
    \includegraphics[width=0.3\textwidth]{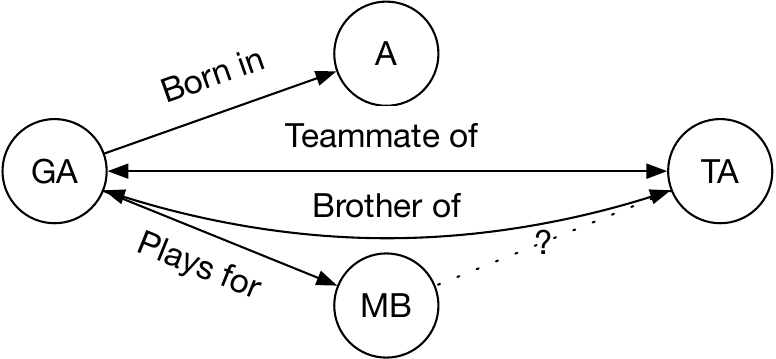}
    \caption{A sample knowledge graph.}
    \vspace{-.15in}
    \label{fig:kg-example}
\end{figure}
\newparagraph{Graphs with Multiple Edge Types} 
We focus on graphs with multiple edge types defined as $G=(V, R, E)$ where $V$ is the set of nodes, $R$ is a set of edge-types or \textit{relations}, and $E$ is the set of edges. Each edge $e = (s, r, d) \in E$ is defined as a triplet containing a source node, relation, and destination node. An example of such a graph is a \emph{knowledge graph}, e.g., Freebase~\cite{freebase}. Here, the source node in a triplet defines a subject (an entity), the relation defines a predicate, and the destination node an object (see example in Figure~\ref{fig:kg-example}). Knowledge graphs are commonly used both in industry and academia to represent real-world facts.

\newparagraph{Graph Embedding Models} 
A graph embedding is a fixed length vector representation for each node (and/or edge-type) in a graph. That is,
each node and relation is represented by a corresponding $d$-dimensional vector $\theta$, also known as an \textit{embedding}~\cite{node2vec}. There are $d(|V| + |R|)$ total learnable parameters. 
To learn these vector representations, embedding models rely on {\em score functions} that capture structural properties of the graph. We denote the score function $f(\theta_s, \theta_r, \theta_d)$ where $\theta_s, \theta_r, \theta_d$ are the vector representations of the elements of a triplet $e = (s, r, d)$. For example, a score function can be the scaled dot product $f(\theta_s, \theta_r, \theta_d) = \theta_s^T\text{diag}(\theta_r)\theta_d$ with the requirement that the parameter vectors are such that $f(\theta_s, \theta_r, \theta_d) \approx 1.0$ if nodes $s$ and $d$ are connected via an edge of type $r$ and $f(\theta_s, \theta_r, \theta_d) \approx 0.0$ otherwise. There are several score functions proposed in the literature ranging from linear score functions~\cite{transE,transR} to dot products~\cite{distmult,complex,rescale} and complex models~\cite{randomwalks,node2vec}.

Score functions are used to form loss functions for training. The goal is to maximize $f(\theta_s, \theta_r, \theta_d)$ if $e \in E$ and minimize it if $e \not\in E$. Triplets that are not present in $E$ are known as \textit{negative edges}. A standard approach~\cite{zheng2020dglke,pytorchbiggraph} is to use the score function $f(\theta_s, \theta_r, \theta_d)$ with a contrastive loss of the form: 
\begin{equation} \label{eq:2}\small
    \mathcal{L} = -\sum\limits_{s, r, d \in E}(f(\mathbf{e_{\theta}}) - \log(\sum\limits_{s', r', d' \not\in E} e^{f(\mathbf{e_{\theta}}')}))
\end{equation}
where $\mathbf{e_{\theta}} = (\theta_s, \theta_r, \theta_d)$ and $\mathbf{e_{\theta}}' = (\theta_s', \theta_r', \theta_d')$.

The first summation term is over all true edges in the graph and the second summation is over all negative edges. There are a total of $|V|^2|R| - |E|$ negative edges in a knowledge graph; this makes it computationally infeasible to perform the full summation and thus it is commonly approximated by \textit{negative sampling}, in which a set of negatives edges is generated by taking a (typically uniform) sample of nodes from the graph for each positive edge. With negative sampling the term in the logarithm is approximated as $\sum\limits_{s, r, d' \in N_e} e^{f(\mathbf{e_{\theta}}')}$. Where $N_e$ is the set of negative samples for $e$.

Graph embeddings are commonly used for \emph{link prediction}, where the similarity of two node vector representations is used to infer the existence of a missing edge in a graph. For example, in the knowledge graph in Figure~\ref{fig:kg-example} we can use the vector representation of $TA$ and $MB$ and the relation embedding for \textit{plays-for} to predict the existence of the edge $TA \xRightarrow{\textit{plays-for}} MB$, marked with a questionmark in the figure.

\newparagraph{The Need for Scalable Training}
The largest publicly available multi-relation graphs have hundreds of millions of nodes and tens of thousands of relations~\cite{wikidata} (Table~\ref{tab:datasets}). Companies have internal datasets which are an order of magnitude larger than these, e.g., Facebook has over 3 billion users~\cite{bronson2013tao}. Learning a 400-dimensional embedding for each of the users will require the ability to store and access $5$ TB of embedding parameters efficiently, far exceeding the CPU memory capacity of the largest machines. Furthermore, using a larger embedding dimension has been shown to improve overall performance on downstream tasks~\cite{Seshadhri5631}. For these two reasons it is important that a system for learning graph embeddings can scale beyond the limitations of GPU and CPU memory.

\newpage

\newparagraph{Scaling Beyond GPU-Memory} We review approaches for scaling the training of graph embedding models out of GPU memory. Prior works follow in two categories: 1) methods that use CPU Memory to store embedding parameters, and 2) methods that use block storage and partitioning of the model parameters. We discuss these two approaches in turn.

Following the first approach, systems such as DGL-KE~\cite{zheng2020dglke} and GraphVite~\cite{zhu2019graphvite}, store node embedding parameters in CPU memory and relation embedding parameters in GPU memory. Shown in Algorithm \ref{alg:SyncTraining}, mini-batch training is performed synchronously and batches are formed and transferred on-demand. While synchronous training is beneficial for convergence, it is resource inefficient. The GPU will be idle while waiting for the batch to be formed and transferred; furthermore, gradient updates also need to be transferred from the GPU to CPU memory and applied to the embedding table, adding additional delays. The effect of this approach on utilization can be seen in Figure~\ref{fig:utilization}, where DGL-KE on average only utilizes about 10\% of the GPU. This approach is also fundamentally limited by the size of the CPU memory, preventing the training of large graph embedding models.

\begin{algorithm}
  \SetAlgoLined
  \nonl\For{$i$ in range$(num\_batches)$}{
    $\mathbf{B_i} = $ getBatchEdges$(i)$\; 
    $\mathbf{\Theta_n} =$ getCpuParameters$(\mathbf{B_i})$\;
    transferBatchToDevice$(B_i, \mathbf{\Theta_n})$\;
    $\mathbf{\Theta_r} =$ getGpuParameters$(\mathbf{B_i})$\;
    $\mathbf{B_\theta} =$ formBatch$(\mathbf{B_i}, \mathbf{\Theta_n}, \mathbf{\Theta_r})$\;
    $\mathbf{G_n}, \mathbf{G_r} =$ computeGradients$(\mathbf{B_\theta})$\;
    updateGpuParameters$(\mathbf{B_i}, \mathbf{G_r})$\;
    transferGradientsToHost$(\mathbf{G_n})$\;
    updateCpuParameters$(\mathbf{B_i}, \mathbf{G_n})$\;
  }
  \caption{Synchronous Embedding Training}
  \label{alg:SyncTraining}
\end{algorithm}

\begin{figure}
    \centering
    \includegraphics[width=0.4\textwidth]{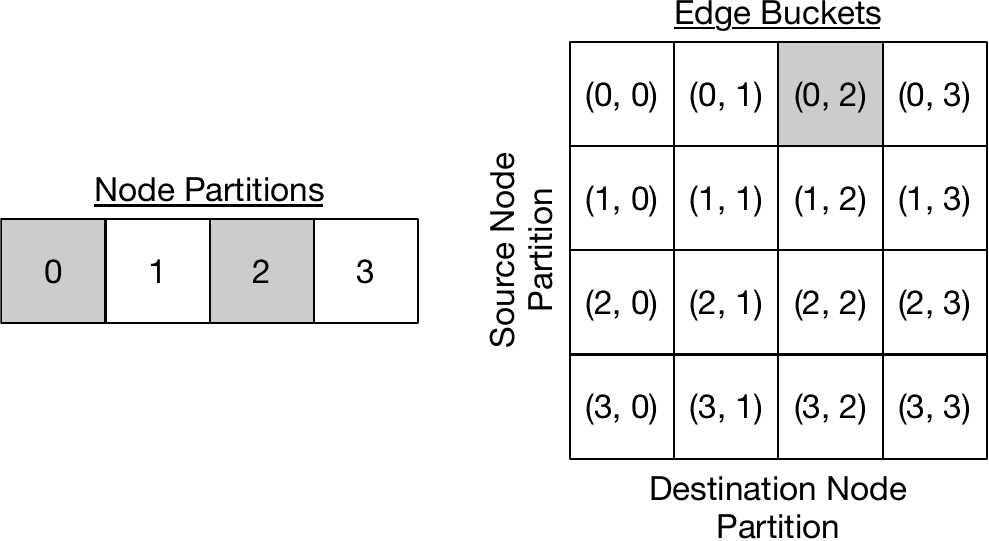}
    \caption{Partitions and edge-buckets with $p=4$. All edges in edge-bucket (0, 2) have a source node in node-partition 0 and a destination node in node-partition 2.}
    \vspace{-.15in}
    \label{fig:partitioning}
\end{figure}

The second approach is adopted by PyTorch BigGraph (PBG)~\cite{pytorchbiggraph}. PBG uses uniform partitioning to split up node embedding parameters into $p$ disjoint partitions and stores them on a block storage device (see example in Figure~\ref{fig:partitioning}). Edges are then grouped according the partition of their source and destination nodes into $p^2$ edge buckets, where all edges in edge bucket $(i, j)$ have a source node which has an embedding in the $i$-th partition, and the destination node which has an embedding in the $j$-th partition. A single epoch of training requires performing mini-batch training over all edge buckets while swapping corresponding pairs of node embedding partitions into memory for each edge bucket. This approach enables scaling beyond CPU memory capacity. 

The major drawback of partitioning is that partition swaps are expensive and lead to the GPU being idle while a swap is happening. In fact, utilization goes towards zero during swaps as shown in Figure~\ref{fig:utilization}. We find that PBG yields an average GPU utilization of 28\%. To best utilize resources, a system using partitioning to scale beyond the memory size of a machine, will need to mitigate overheads that arise from swapping partitions.

\subsection{Data Movement Challenges}
We discuss how to optimize data movement and related challenges that Marius' architecture addresses; we discuss the architecture in detail in Sections~\ref{pipelined_training} and~\ref{sec:out-of-memory}.

\begin{figure*}[t!]
    \centering
    \includegraphics[width=\textwidth]{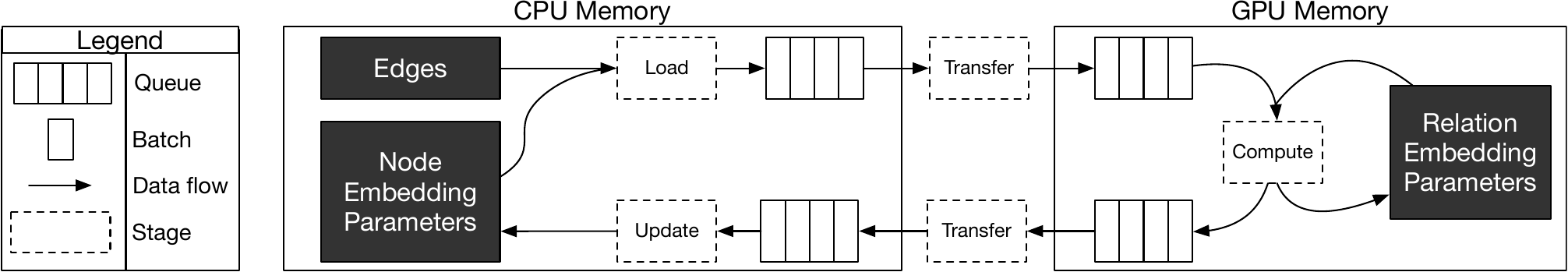}
    \caption{Marius training pipeline.}
    \vspace{-.1in}
    \label{fig:MariusPipeline}
\end{figure*}

\newparagraph{Traditional Optimizations for Data Movement}
Pipelining is a common approach used in a number of system designs to overlap computation with data movement, thereby improving utilization~\cite{hennessy2011computer,narayanan2019pipedream,palkar2019optimizing}. Using an image classifier as an example, a simple pipeline will consist of multiple worker threads that pre-process training images in parallel, forming batches and transferring them to the GPU. Once on the GPU, batches of training data are pushed onto a queue, with a training process constantly polling the queue for new batches. By keeping the queue populated with new batches, the GPU will be well utilized. 

In IO-bound applications, buffer management can also be used to prevent unnecessary IO by caching data in memory. Buffer management is well studied in the area of databases and operating systems and has been applied to a myriad of applications and workloads~\cite{ramakrishnan2003database,hellerstein2007architecture}. When using a buffer, the order in which data is accessed and swapped impacts end-to-end performance. When the data access pattern exhibits good locality, buffer managers typically yield good performance. Additionally, if the ordering is known ahead of time the buffer manager may prefetch data items and use Belady's optimal cache replacement algorithm to minimize IO~\cite{belady1966study}. 

In graph processing, locality-aware data layouts of graph edges have been shown to improve locality of accesses and performance of common graph algorithms such as PageRank~\cite{mcsherry2015scalability}. One such data layout, utilizes Hilbert space filling curves to define an ordering over the adjacency matrix of the graph. The ordering produced is a 1D index that preserves the locality of the 2D adjacency matrix. Storing and accessing edges according to this index improves OS cache hit rates~\cite{mcsherry2015scalability,mosaic}.

\newparagraph{Challenges for Graph Embeddings}
For large graphs and embedding sizes, graph embedding models can be multiple orders of magnitude larger than the GPU's memory capacity, a key difference from deep neural network models that typically fit in a single GPU. To design a pipeline for graph embedding training, not only will training data (formed by considering edges) have to be piped to the GPU but also the corresponding model parameters (the node and relation embeddings of the end-points and the type of each edge). Furthermore, model updates need to be piped back from the GPU and applied to the underlying storage. By pipelining model parameters and updates, we introduce the possibility of stale parameters, which must mitigated (see Section~\ref{pipelined_training}). 

Buffer management techniques paired with data orderings can be used to buffer partitions in CPU memory to reduce IO from disk. However, we find that prior locality-aware data orderings such as space-filling curves fall short and still result in IO bound training due to a non-optimal amount of swaps (Sections~\ref{subsec:edgeBucketOrderings} and~\ref{subsec:partition_ordering}). To address this challenge we propose a buffer-aware data ordering which results in a near-optimal number of swaps, referred to as {\em BETA ordering}, in Section~\ref{sec:out-of-memory}.

\section{Pipelined Training Architecture} \label{pipelined_training}

We review Marius' pipelined architecture for training graph embedding models. We first discuss the overall design, then the details of each stage, and finally discuss how staleness arises due to interleaving computation with data movement and how we can mitigate it.

\newparagraph{Pipeline Design} \label{pipeline_design}
Our architecture follows Algorithm~\ref{alg:SyncTraining} and divides its steps into a five-stage pipeline with queues separating each stage (Figure~\ref{fig:MariusPipeline}). Four stages are responsible for data movement operations, and one for model computation and in-GPU parameter updates. The four data movement stages have a configurable number of worker threads, while the model computation stage only uses a single worker to ensure that relation embeddings stored on the GPU are updated synchronously. 

We now describe the different stages of the pipeline and draw connections to the steps in Algorithm~\ref{alg:SyncTraining}:

\vspace{2pt}\noindent\textbf{Stage 1: Load} This stage is responsible for loading the edges (i.e., entries that correspond to a pair of node-ids and the type of edge that connects them) and the corresponding node embedding vectors that form a batch of inputs used for training. The edge payload constructed in this stage includes the true edges appearing in the graph and a uniform sample of negative edges (i.e., fake edges) necessary to form the loss function in Equation~\ref{eq:2} (Lines 1-2 in Algorithm~\ref{alg:SyncTraining}).

\vspace{2pt}\noindent\textbf{Stage 2: Transfer} The input to this stage consists of the edges (node-id and edge-type triples) and the node embeddings from the previous stage. Worker threads in this stage asynchronously transfer data from CPU to GPU using \texttt{cudaMemCpy} (Line 3 in Algorithm~\ref{alg:SyncTraining}).

\vspace{2pt}\noindent\textbf{Stage 3: Compute} The compute stage is the only stage that does not involve data movement. This stage takes place on GPU where the payload of edges and node embeddings created in Stage 1 is combined with relation embedding vectors (corresponding to the edge-type associated with each entry) to form a full batch. The worker thread then computes model updates and applies updates to relation embeddings stored in the GPU. The updates to node embeddings (i.e., the scaled gradients that need to be added to the previous version of the node embedding parameters) are placed on the output queue to be transferred from GPU memory (Lines 4-7 in Algorithm~\ref{alg:SyncTraining}).

\vspace{2pt}\noindent\textbf{Stage 4: Transfer} The node embedding updates are transferred back to the CPU. We use similar mechanisms as in Stage 2 (Line 8 in Algorithm~\ref{alg:SyncTraining}).

\vspace{2pt}\noindent\textbf{Stage 5: Update.} The final stage in our pipeline applies node embedding updates to stored parameters in CPU memory (Line 9 in Algorithm~\ref{alg:SyncTraining}).  

This hybrid-memory architecture allows us to execute sparse parameter updates asynchronously (i.e., the node embedding parameter updates) and dense updates (i.e., the relation embedding parameter updates) synchronously, and optimize resource utilization as we show experimentally in Section~\ref{sec:experiments}.

\newparagraph{Bounded Staleness} \label{bounded_staleness}
The main challenge with using a pipelined design as described above, is that it introduces \emph{staleness} due to asynchronous processing \cite{10.5555/2999611.2999748}. To illustrate this, consider a batch entering the pipeline (Stage 1) with the embedding for node $A$. Once this batch reaches the GPU (Stage 3), the gradients for the embedding for $A$ will be computed. While the gradient is being computed, consider another batch that also contains the embedding for node $A$ entering the pipeline (Stage 1). Now, while the updates from the first batch are being transferred back to the CPU and applied to parameter storage, the second batch has already entered the pipeline, and thus it contains a stale version of the embedding for node $A$.  

To limit this staleness, we bound the number of batches in the pipeline at any given time. For example, if the bound is $4$, embeddings in the pipeline will be at worst $4$ updates behind. However, due to the sparsity of node embedding updates, it is unlikely a node embedding will even become stale. To give a realistic example, take the Freebase86m graph which has 86 million nodes. A typical batch size and staleness bound for this benchmark is 10,000 and 16 respectively. Each batch of 10,000 edges will have at most 20,000 node embeddings and given this staleness bound there can be at most 320,000 node embeddings in the pipeline at any given time, which is just about .4\% of all node embeddings. Even with this worst case, only a very small fraction of node embeddings will be operated on at a given time. The same property does not hold for relation embeddings since there are very few of them (15K in Freebase86m), hence our design decision to keep relation embeddings in GPU memory and update them synchronously, bypasses the issue of stale relation embeddings.
We study the effect of staleness and Marius' performance as we vary the bound in Section~\ref{sec:exp_microbenchmarks}.

\section{Out-of-memory Training}
\label{sec:out-of-memory}

As described in Section~\ref{sec:background}, to learn embedding models for graphs that do not fit in CPU memory, existing systems partition the graph into non-overlapping blocks. They correspondingly partition the parameters as well so that they can be loaded sequentially for processing.
However as IO from disk can be slow (e.g, a partition can be around 10s of GB in size), it is desirable to hide the IO wait times and minimize the number of swaps from disk to memory. 
In this section, we describe how we can effectively hide IO wait time by integrating our training pipeline with a \emph{partition buffer} that constitutes an in-memory buffer of partitions. We also describe how we can minimize the number of swaps from disk to memory by developing a new ordering for traversing graph data.

\newparagraph{Partition-based training}
Consider a graph that is partitioned into $p^2$ edge buckets corresponding to $p$ node-partitions. Training one epoch requires iterating over all $p^2$ edge buckets, where each edge in a given bucket $(i, j)$, will have a source node in partition $i$ and destination node in partition $j$. 

When processing an edge bucket $(i, j)$, node partition $i$ and node partition $j$ must be present in the CPU partition buffer in order for learning to proceed using the pipelined training architecture (see Section~\ref{pipeline_design}). If either one is not present, it must be loaded from disk and \textit{swapped} into the buffer, replacing an already present partition if the buffer is full. Partition-based training is described in Algorithm~\ref{alg:EdgeBucket}.

Given a partitioned graph, there are a number of edge bucket orderings that can be used for traversal. To minimize the number of times partitions need to be loaded from disk, \textit{we seek an ordering over edge buckets which minimizes the number of required partition swaps}.

We note that once an edge bucket ordering has been selected, we can further mitigate IO overhead by 1) prefetching to load node partitions as they are needed in the near future and 2) using the optimal buffer eviction policy which removes partitions used farthest in the future. 

We next discuss the problem of determining an optimal ordering over edge buckets and describe the \emph{BETA ordering}, a new ordering scheme that achieves near-optimal number of partition swaps.

\begin{algorithm}[t]
    \SetInd{.25em}{1em}
  \SetAlgoLined
  Buffer $= \{\}$\;
  \For{$k$ in range$(p^2)$}{
        $\mathbf{E_{ij}}, i, j =$ getEdgeBucket(Ordering$[k]$)\;
        \If{i not in Buffer}{
            \If{Buffer.size() == $c$} {
                Buffer.evictFurthest(Ordering, $k$)\;
            }
            Buffer.admit$(i)$\;
        }
        \If{j not in Buffer}{
            \If{Buffer.size() == $c$} {
                Buffer.evictFurthest(Ordering, $k$)\;
            }
            Buffer.admit$(j)$\;
        }
        $\mathbf{\Theta_{i}}$ = Buffer.get$(i)$; // Source Node Partition \\
        $\mathbf{\Theta_{j}}$ = Buffer.get$(j)$; // Destination Node Partition \\
        trainEdgeBucket$(\mathbf{E_{ij}}, \mathbf{\Theta_{i}}, \mathbf{\Theta_{j}})$\;
  }
  \caption{Training Using a Partition Buffer}
  \label{alg:EdgeBucket}
\end{algorithm}

\subsection{Edge Bucket Orderings}
\label{subsec:edgeBucketOrderings}
We develop an edge bucket ordering scheme that minimizes the number of swaps. First, we derive a lower bound on the number of swaps necessary to complete one training epoch for a buffer of size $c$ and $p~(p >= c)$ partitions.
To derive the lower bound, \emph{we view an edge bucket ordering as a sequence of partition buffers} over time, where each item in the sequence describes what node partitions are in the buffer at that point. Each successive buffer differs by one swapped partition. 

Given such a sequence, an edge bucket ordering can be constructed by processing edge bucket $(i, j)$ when partitions $i$ and $j$ are in the buffer. For simplicity, we can do this the first time $i$ and $j$ appear together. Note that 1) $i$ and $j$ must appear together at least once otherwise no ordering over all edge buckets can be constructed, 2) self-edge buckets (i.e. $(i, i)$) can also be added to the ordering the first time $i$ appears in the buffer, 
and 3) there are many edge bucket orderings with the same sequence of partition buffers (depending on the order in which edge buckets in a particular buffer are processed). Viewed in this light, \textit{we seek the shortest (min. swaps) buffer sequence where all node partition pairs appear together in the buffer at least once}. 

\begin{algorithm}[!t]
    \SetInd{.25em}{1em}
    \SetAlgoLined
    PartitionBufferSequence $= \{\}$\;
    CurrentBuffer $= [0 \dots c-1]$\;
    OnDisk $= [c \dots p-1]$\;
    PartitionBufferSequence.append$($CurrentBuffer$)$\;
    
    \While{OnDisk.size$() > 0$}{
        \For{$i$ in range$($OnDisk.size$())$}{
            swap$($CurrentBuffer$[-1]$, OnDisk$[i])$\;
            PartitionBufferSequence.append$($CurrentBuffer$)$\;
        }
        $n = 0$\;
        \For{$i$ in range$(c-1)$}{
            \If{$i \ge$ OnDisk.size$()$}{
                \textbf{break}\;
            }
            $n = n+1$\;
            CurrentBuffer$[i]$ $=$ OnDisk$[i]$\;
            PartitionBufferSequence.append$($CurrentBuffer$)$\;
        }
        OnDisk $=$ OnDisk$[n:end]$\;
    }
    \Return PartitionBufferSequence\;
    \caption{\textit{BETA} Ordering Buffer Sequence}
    \label{alg:BETA}
\end{algorithm}
\begin{algorithm}[!t]
    \SetInd{.25em}{1em}
    \SetAlgoLined
    EdgeBuckets $= \{\}$\;
    SeenPairs $=$ zeros$(p, p)$\;
    \For{Buffer in PartitionBufferSequence}{
        NewEdgeBuckets $= \{\}$\;
        \For{$i$ in Buffer}{
            \For{$j$ in Buffer}{
                \If{SeenPairs$[i, j] == 0$}{
                    SeenPairs$[i, j] = 1$\;
                    NewEdgeBuckets.append$((i, j))$\;
                }
            }
        }
        shuffle$($NewEdgeBuckets$)$\;
        EdgeBuckets.append$($NewEdgeBuckets$)$\;
    }
    \Return EdgeBuckets\;
    \caption{Buffer Seq. to Edge Bucket Order}
    \label{alg:buffer_to_bucket}
\end{algorithm}

\newparagraph{Lower bound} We assume that initializing the first full buffer does not count as part of the total number of swaps as all orderings must incur this cost. Thus, there are $\frac{p(p-1)}{2}$ (the total number of pairs) minus $\frac{c(c-1)}{2}$ (the number of pairs we get in the first buffer) remaining partition pairs that must appear together in the partition buffer. On any given swap, the most new pairs we can cover is if the partition entering the buffer has not been paired with anything already in the buffer (everything in the buffer has already been paired with everything else in the buffer). Thus, for each swap the best we can hope for is to get $c-1$ pairs we have not already seen. With this in mind a lower bound on the minimum number of swaps required is:

\begin{equation}
    \left\lceil \frac{\frac{p(p-1)}{2} - \frac{c(c-1)}{2}}{c-1}\right\rceil
\end{equation}

We use this lower bound to evaluate the performance of different edge bucket orderings in the next section. We experimentally show that the new ordering strategy we propose is nearly optimal with respect to this bound.

\begin{figure*}
\centering
\includegraphics[width=0.9\textwidth]{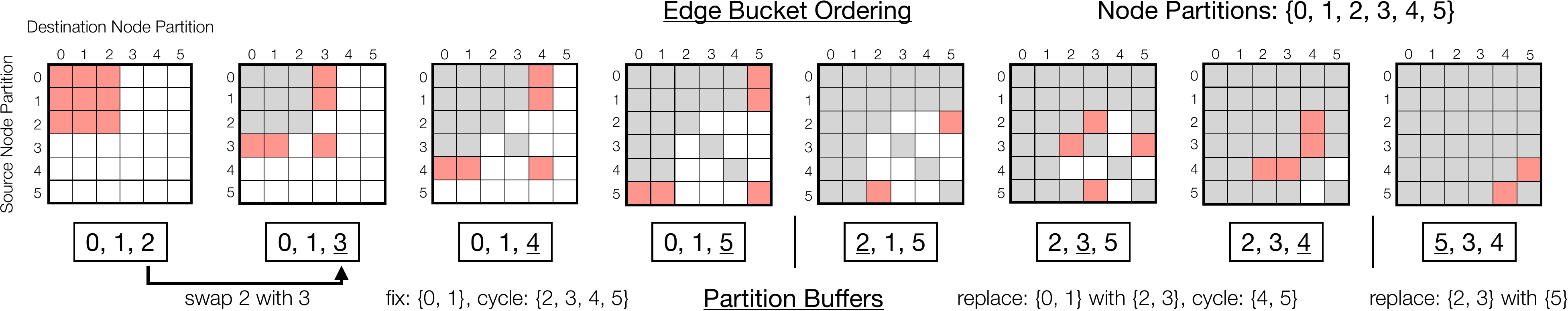}
\caption{Example \textit{BETA} ordering for $p=6$ and $c=3$. The sequence of partition buffers corresponds to first fixing \{0, 1\}, then replacing \{0, 1\} with \{2, 3\}, fixing \{2, 3\}, and finally replacing \{2, 3\} with \{5\}. Each successive buffer differs by one swap. A corresponding edge bucket ordering is shown above the buffers. For each partition buffer in the sequence, all previously unprocessed edge buckets which have their source and destination node partitions in the buffer are added to the ordering (red edge buckets). For each buffer, these edge buckets can be added in any order.}
\vspace{-0.1in}
\label{fig:BETA_example}
\end{figure*}

\begin{figure}[!t]
\centering
\begin{subfigure}{.42\columnwidth}
  \centering
  \includegraphics[width=\linewidth]{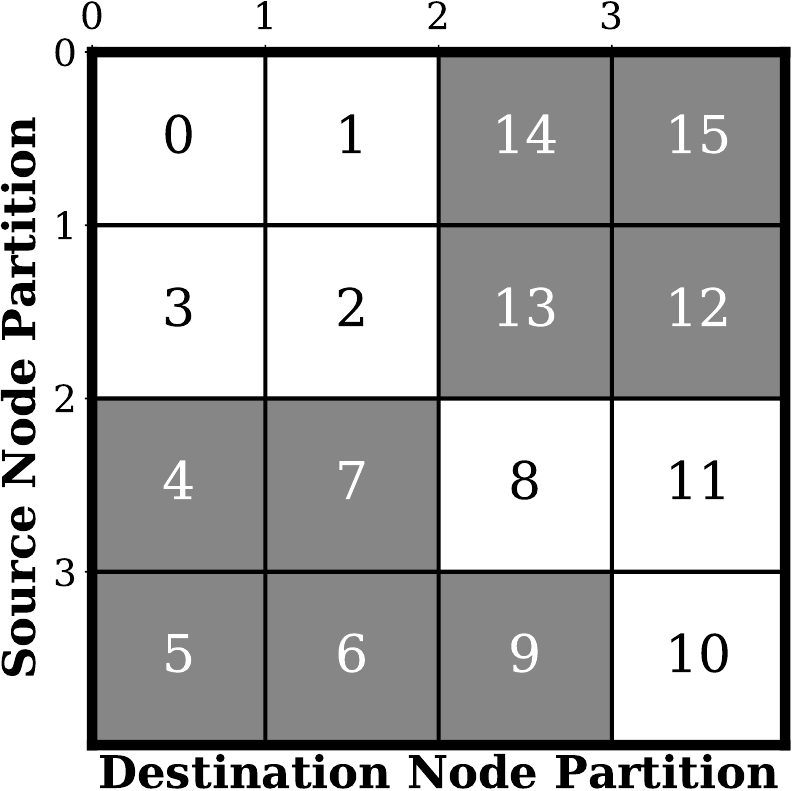}
  \caption{Hilbert Ordering}
  \label{fig:sub1}
\end{subfigure}%
\hspace{5pt}
\begin{subfigure}{.42\columnwidth}
  \centering
  \includegraphics[width=\linewidth]{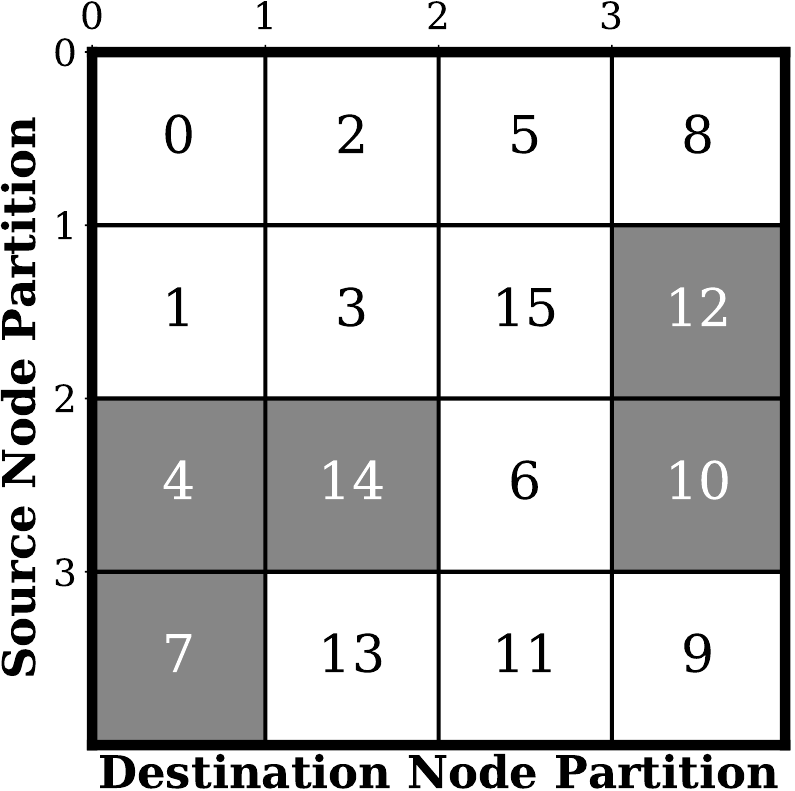}
  \caption{\textit{BETA} Ordering}
  \label{fig:sub2}
\end{subfigure}
\caption{Hilbert and \textit{BETA} edge bucket orderings. Numbers indicate the order in which the bucket is processed. Gray cells indicate misses to the buffer.}
\vspace{-0.1in}
\label{fig:edge_bucket_orderings}
\end{figure}

\newparagraph{\textit{BETA} ordering}
We describe the \emph{Buffer-aware Edge Traversal Algorithm (BETA)}, an algorithm to compute the edge bucket ordering that achieves close to optimal number of partition swaps and improves upon locality-aware orderings such as Hilbert space-filling curves~\cite{hilbert1891ueber}.

Algorithm~\ref{alg:BETA} describes how the \textit{BETA} ordering of partition buffers is generated. Consider a partition buffer that was initialized with the first $c$ node-partitions in the graph (Line 2 in Algorithm~\ref{alg:BETA}). The remaining $p-c$ node-partitions start on disk (Line 3 in Algorithm~\ref{alg:BETA}). To generate the partition buffer sequence we then proceed as follows: First we fix the leading $c-1$ node-partitions in the buffer and swap each of the outstanding partitions into the final buffer spot, one at a time (Line 6-8 in Algorithm~\ref{alg:BETA}). Each swap creates a new partition buffer in the sequence. Once this is complete, the fixed $c-1$ partitions have been paired in the buffer with all other node-partitions and are therefore no longer needed. We refresh our buffer by replacing the finished $c-1$ partitions with new node-partitions from the unfinished set on disk (Line 10-15 in Algorithm~\ref{alg:BETA}). The incoming partitions can then be deleted from the on disk set (Line 16 in Algorithm~\ref{alg:BETA}) since they are now in the buffer. As before, each swap results in a partition buffer added to the sequence. 
We repeat this process until there are no remaining unfinished node-partitions (Line 5 and 11-12 in Algorithm~\ref{alg:BETA}). As described at the beginning of Section~\ref{subsec:edgeBucketOrderings} and in more detail in Algorithm~\ref{alg:buffer_to_bucket}, the partition buffer sequence can be easily converted to the final edge bucket ordering. We show an example \textit{BETA} ordering in Figure~\ref{fig:BETA_example}. 

We observe that our \textit{BETA} ordering has a number of useful properties that make it advantageous to implement in practice. Since all partitions are symmetrically processed we do not need to track any extra state or use any priority mechanisms. Further, for every disk IO (swap) with a fixed set of $c-1$ partitions (Line 7 in Algorithm~\ref{alg:BETA}), the incoming node-partition has yet to be paired with any other partition in the buffer. This means we can process $c-1$ edge buckets before performing another swap---the most possible (excluding self edge buckets)---allowing us to hide IO operations behind longer compute times. The only bottleneck arises when the fixed $c-1$ partitions are replaced, but this only happens at most $\left\lfloor \frac{p-c}{c-1} \right\rfloor + 1$ times in one epoch. Additionally, the \textit{BETA} ordering can be randomized to create different graph traversals by shuffling which partitions start in the buffer, by permuting the buffer and/or on disk set before Line 6 in Algorithm~\ref{alg:BETA}, or by permuting the on disk set before Line 10 in Algorithm~\ref{alg:BETA}.

Finally, we analyze the number of swaps generated by the \textit{BETA} ordering: given $p$ partitions and a buffer of size $c$ the number of swaps is 
\begin{align}
\begin{split}
	(p-c) + (x+1)&\left[(p-c) - \frac{1}{2} x (c-1) \right] \\ \text{where } x = &\left\lfloor \frac{p-c}{c-1} \right\rfloor.
\end{split}
\end{align}

\newparagraph{Comparison with Hilbert, lower bound}
We compare the number of IO operations incurred by the \textit{BETA} ordering with space-filling curve based orderings, and the analytical lower bound. Space filling curve orderings like Hilbert~\cite{hilbert1891ueber} attempt to define a graph traversal that preserves 2D locality over the $n \times n$ matrix of edge buckets. We also compare to a second version of the Hilbert ordering, termed Hilbert Symmetric, which modifies the former by processing edge buckets $(i,j)$ and $(j,i)$ successively. A key advantage of the \textit{BETA} ordering when compared to these methods is that it is buffer-aware, i.e., the algorithm knows the buffer size and specifically aims to minimize partition swaps. In contrast, space-filling curve based orderings are unaware of this information, aiming instead for locality.

\begin{figure}[!t]
   \centering
    \includegraphics[width=.9\columnwidth]{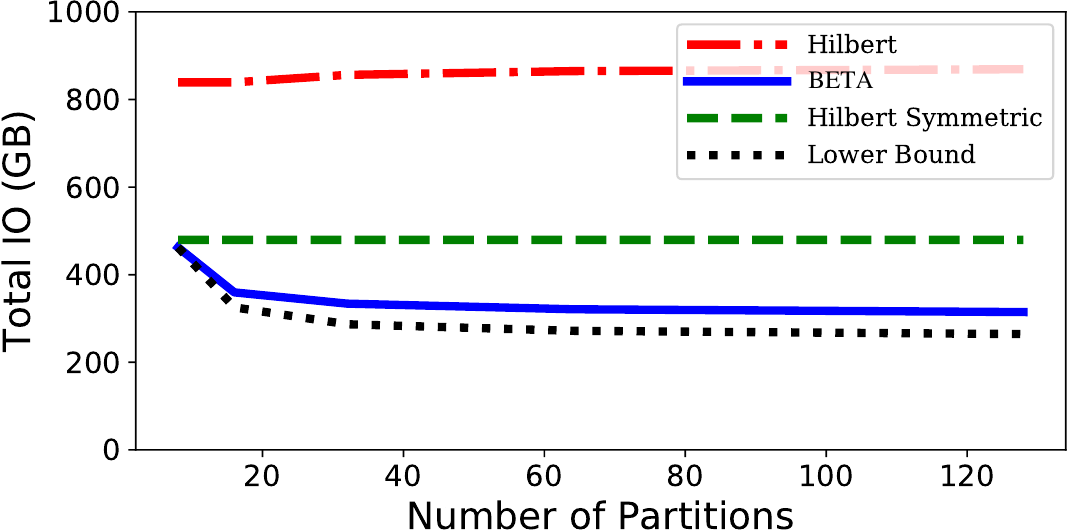}
    \caption{Simulated total IO performed during a single epoch of training Freebase86m with $d=100$.}
    \vspace{-0.1in}
    \label{fig:greedy-io}
\end{figure}

We illustrate how the \textit{BETA} ordering compares to a Hilbert space-filling curve on a small $p=4, c=2$ case in Figure~\ref{fig:edge_bucket_orderings}. We see that while the Hilbert ordering has nine buffer misses the \textit{BETA} ordering only has five misses. We also performed simulations to compare each method. Figure~\ref{fig:greedy-io} shows the number of IO accesses when varying $p$ and using a buffer with size $\frac{p}{4}$ for the \textit{BETA}, Hilbert, and Hilbert Symmetric orderings, together with the lower bound. The \textit{BETA} ordering yields nearly optimal performance across partition configurations and requires significantly less IO than the other methods.

We leave an investigation of a provably-optimal ordering for future work. Our initial studies have shown that there exist cases of $p$, $c$ where no valid ordering can match the lower bound as well as cases where an ordering which requires slightly fewer swaps than the \textit{BETA} ordering does exist. Thus, the optimal algorithm requires IO-swaps somewhere between the lower bound and the \textit{BETA} ordering in Figure~\ref{fig:greedy-io}.

\subsection{Partition Buffer} \label{subsec:partition_buffer}
We next describe mechanisms that we use in the partition buffer to further minimize IO overhead.
The partition buffer is a fixed sized memory region that has capacity to store $c$ embedding partitions in memory. We \emph{co-design} the buffer replacement policies with the \textit{BETA} ordering described above. Co-designing the edge traversal with buffer replacement policy means that we have knowledge about which partitions will be accessed in the future. This allows the buffer to use the optimal replacement policy: \textit{evict the partition that will be used furthest in the future}~\cite{belady1966study}. Given this policy we also design a prefetching mechanism that can minimize the amount of time spent waiting for partitions to swap. Again, based on knowing the order in which partitions are used, we use a prefetching thread that reads the next partition in the background. Correspondingly when a partition needs to be evicted from memory, we perform asynchronous writes using a background writer thread.

\begin{table*}[t]\scriptsize
\centering
\setlength\tabcolsep{4pt}
\begin{tabular}{ |c|c|c|c|c|c|p{0.57\textwidth}| } 
\hline
Name & Type & |E| & |V| & |R| & Size & Hyperparameters \\ 
\hline
\hline
FB15k & KG & 592k & 15k & 1.3k & 52 MB & \small$d=400$, $lr=.1$, $b=10^4$, $n_t=10^3$, $\alpha_{n_t}=.5$, FilteredMRR \\ 
\hline
LiveJournal & Social & 68M & 4.8M & - & 1.9 GB & \small$d=100$, $lr=.1$, $b=5 \times 10^4$, $n_t=10^3$, $\alpha_{n_t}=.5$, $n_e=10^4$, $\alpha_{n_e}=0$ \\ 
\hline
Twitter & Social & 1.46B & 41.6M & - & 33.2 GB & \small$d=100$, $lr=.1$, $b=5 \times 10^4$, $n_t=10^3$, $\alpha_{n_t}=.5$, $n_e=10^3$, $\alpha_{n_e}=.5$ \\ 
\hline
Freebase86m & KG & 338M & 86.1M & 14.8K & 68.8 GB & \small$d=100$, $lr=.1$, $b=5 \times 10^4$, $n_t=10^3$, $\alpha_{n_t}=.5$, $n_e=10^3$, $\alpha_{n_e}=.5$\\ 
\hline
\end{tabular}
\caption{Datasets used for evaluation. The size column indicates total size of embedding parameters with the embedding dimension $d$, including the Adagrad optimizer state. $lr$: learning rate, $b$: batch size, $n_t$: training negatives, $\alpha_{n_t}$: train degree-based negatives fraction, $n_e$: evaluation negatives, $\alpha_{n_e}$: eval degree-based negatives fraction.}
\vspace{-.15in}
\label{tab:datasets}
\end{table*}

\section{Evaluation}
\label{sec:experiments}

We evaluate Marius on standard benchmarks using a single AWS P3.2xLarge instance and compare against SoTA graph embedding systems. We show that:

\vspace{2pt}\noindent(1) Due to optimized resource utilization, Marius yields up to 10$\times$ faster training in comparison to SoTA systems and cost reductions on cloud resources between 2.9$\times$ and 7.5$\times$ depending on the configuration.

\vspace{2pt}\noindent(2) The \textit{BETA} ordering reduces IO required by up to 2$\times$ when compared to other locality-based graph orderings, thus alleviating the IO bottleneck during training.

\vspace{2pt}\noindent(3) Marius is able to scale to graph embedding models that rely on increased vector dimensions to achieve higher accuracy. Due to the increased vector dimensions these models exceed CPU memory size. For instance, we show that Marius can learn an embedding model using 800-d vector representations on a graph with 86M nodes on a single machine. In this configuration there are 550 GB of total parameters and optimizer state, which is 35$\times$ GPU memory size and 9$\times$ CPU memory size.

\subsection{Setup}
\label{subsec:experiment_setup}

\newparagraph{Implementation} Marius is implemented in about 10,000 lines of C++. We use LibTorch~\cite{libtorch}, which is the C++ API of PyTorch, as the underlying tensor engine. LibTorch provides access to the wide-ranging functionality of PyTorch, making it easy to extend Marius to support more complex embedding models. We also implement an abstracted storage API, which allows for embedding parameters to be stored and accessed across a variety of backends under one unified API. This allows us to easily switch between storage backends, say from using a CPU memory-based backend to a disk-based backend.

\newparagraph{Hardware Setup} Single machine experiments are run on a single AWS P3.2xLarge instance which has: 1 Telsa V100 GPU with 16 GB of memory, 8 vCPUs with 64 GB of memory, and an attached EBS volume with 400 MBps of read and write bandwidth. For multi-GPU experiments, we use the AWS P3.16xLarge instance which has 8 Tesla V100 GPUs with 16 GB of memory each, 64 vCPUs, and 524 GB of CPU memory. For distributed multi-core experiments we use 4 c5a.8xLarge instances with 32 vCPUs and 69 GB of CPU memory each. DGL-KE ran out of memory when using a single GPU with the Twitter and Freebase86m datasets. For these cases, we use a larger machine with 1 Telsa V100 GPU with 32 GB of memory and 200 CPUs with 500 GB of memory.

\newparagraph{Datasets} For our evaluation, we use standard benchmark datasets that include social networks (Twitter~\cite{twitter}, Livejournal~\cite{livejournal}) and knowledge graphs (FB15k and Freebase86m~\cite{pytorchbiggraph,zheng2020dglke} derived from Freebase~\cite{freebase}). A summary of the dataset properties is shown in Table \ref{tab:datasets}. FB15k uses an 80/10/10 train, validation and test split. All others use a 90/5/5 split.

\newparagraph{Embedding Models} 
On FB15k, we use ComplEx~\cite{complex} and DistMult~\cite{distmult}. On LiveJournal and Twitter we use \textit{Dot}~\cite{nrl-tutorial}, which is a dot product between the node embeddings of an edge. On Freebase86m we use ComplEx embeddings. We chose these models to match the evaluation of Zheng et al. ~\cite{zheng2020dglke} and Lerer et al. ~\cite{pytorchbiggraph}. 

\newparagraph{Hyperparameters}
To ensure fair comparisons, we use the same hyperparameters across each system instead of tuning separately. Hyperparameter values for each configuration were chosen based on those used in the evaluation of DGL-KE and PBG and are shown in Table~\ref{tab:datasets}. All systems use the Adagrad optimizer~\cite{adagrad} for training, which empirically yields much higher-quality embeddings over SGD. 
One drawback of using Adagrad is that it effectively requires storing a learning rate per parameter, doubling the overall memory footprint of the embeddings during training. For Marius, we use a staleness bound of 16 for all cases which utilize the pipeline. 

\newparagraph{Evaluation Task and Metrics}
We evaluate the quality of the embeddings using the link prediction task. Link prediction is a commonly used evaluation task in which embeddings are used to predict if a given edge is present in the graph. Link prediction metrics reported are Mean Reciprocal Rank (MRR) and Hits@$k$, which are derived from the rank of the score of each candidate edge, where the scores are produced from the embedding score function $f$. For a given candidate edge $i$, it has a rank $r_i$ which denotes the position of the score of the candidate edge in descending sorted array $S_i$, where $S_i$ contains the score of the candidate edge and the scores of a set of negative samples. Given this, the MRR and Hits@$k$ can be computed from a set of candidate edges $C$ as follows: 
$\dfrac{1}{|C|}\sum_{i \in C}\dfrac{1}{r_i}$ and $\dfrac{1}{|C|}\sum_{i \in C}\mathbb{1}_{r_i <= k}$ respectively.

Metrics can be filtered or unfiltered. Filtered evaluation involves comparing candidate edges with $|N|$ negative samples, produced using all of the nodes in the graph. Some of the produced negative samples will be false negatives, which will not be used in filtered evaluation. Because all nodes in the graph are used, filtered evaluation is expensive for large graphs. Unfiltered evaluation samples $n_e$ nodes from the graph, with a fraction $\alpha_{n_e}n_e$ by degree and $(1 - \alpha_{n_e})n_e$ uniformly. False negatives are not removed in unfiltered evaluation, but will not be common if $n_e<<|V|$. Unfiltered evaluation is much less expensive and is well suited for large scale graphs. We use filtered metrics only on FB15k and unfiltered metrics elsewhere. The same evaluation approach is adopted by prior systems ~\cite{pytorchbiggraph}.

\subsection{Comparison with Existing Systems} \label{sec:system_comparisons}
To demonstrate that Marius utilizes resources better than current SoTA systems leading to faster training, we compare Marius with PBG and DGL-KE on four benchmark datasets. We do not compare with GraphVite since it is significantly slower than DGL-KE as reported in Zheng et al.~\cite{zheng2020dglke}. FB15k and LiveJournal fit in the machine's GPU memory and therefore do not have data movement overheads. Twitter exceeds GPU memory which introduces data movement overheads from storing parameters off-GPU. Freebase86m exceeds the CPU memory of the machine, which prevents DGL-KE from training these embeddings on a single P3.2xLarge instance, therefore we only compare against PBG. 

\newparagraph{FB15k} In this experiment, we compare Marius with PBG and DGL-KE on FB15k to show that Marius achieves similar embedding quality as the other systems on a common benchmark. We measure the FilteredMRR, Hits@k, and runtime of the systems when training ComplEx and DistMult embeddings with $d=400$ to peak accuracy, averaged over five separate runs. Results are shown in Table \ref{tab:1}. It should be noted that all parameters and training data fit in GPU memory for this dataset. We find that Marius achieves near identical metrics as PBG when learning the same embeddings, this is expected as both systems have similar implementations for sampling edges and negative samples. DGL-KE on the other hand only achieves a similar FilteredMRR. DGL-KE has implementation differences for initialization and sampling which likely account for the difference in metrics. While Marius is not designed for small knowledge graphs, we can see that it performs comparably to SoTA systems, achieving similar embedding quality in lesser time.

\begin{table}[t]\scriptsize
\centering
\begin{tabular}{ |c|c|>{\centering\arraybackslash}p{0.38in}|>{\centering\arraybackslash}p{0.2in}|>{\centering\arraybackslash}p{0.2in}|>{\centering\arraybackslash}p{0.5in}| } 
\hline
System & Model & Filtered & \multicolumn{2}{|c|}{Hits} & Time \\ 
& & MRR & @1 & @10 & (s)\\
\hline
\hline
DGL-KE & ComplEx  & .795 & .766 & .848 & 35.6s $\pm$ .69 \\ 
\hline
PBG & ComplEx & .795 & .736 & .888 & 40.3s $\pm$ .1\\ 
\hline
Marius & ComplEx & .795 & .736 & .888 & \textbf{27.7s} $\pm$ .12\\ 
\hline
\hline
DGL-KE & DistMult & .792 & .766 & .848 & 32.8s $\pm$ .88 \\ 
\hline
PBG & DistMult & .790 & .728 & .888 & 46.2s $\pm$ .46\\ 
\hline
Marius & DistMult & .790 & .727 & .889 & \textbf{28.7s} $\pm$ .15 \\ 
\hline
\end{tabular}
\caption{FB15k Results. All systems reach peak accuracy at about the same number of epochs with 30 and 35 epochs for ComplEx and DistMult respectively.}
\vspace{-.1in}
\label{tab:1}
\end{table}

\newparagraph{LiveJournal}
To show that the systems are comparable on social graphs, we compare the quality of $100$-dimensional embeddings learned by the three systems using a dot product score function. While Livejournal is two orders of magnitude larger than FB15k, all parameters still fit in GPU memory with a total of 2 GB. As before, we measure MRR, hits@k, and runtime, averaging over three runs; but we use unfiltered MRR instead of FilteredMRR. We do so because FilteredMRR is computationally expensive to evaluate on larger graphs (Section \ref{subsec:experiment_setup}). Instead of using all nodes in the graph to construct negative samples, we sample 10,000 nodes uniformly for evaluation, as done in Lerer et al.~\cite{pytorchbiggraph}. Results are shown in Table~\ref{tab:2}. We see that all three systems achieve near identical metrics for this dataset. There are slight differences in runtimes that can be attributed to implementation differences. PBG checkpoints parameters after each epoch, while this is optional in Marius and DGL-KE. Without checkpointing, PBG would likely achieve similar runtimes to DGL-KE and Marius. Overall, we find that Marius performs as well or better than SoTA systems on this social graph benchmark. 

\begin{table}[t]\scriptsize
  \centering
  \begin{tabular}{ |c|c|>{\centering\arraybackslash}p{0.38in}|>{\centering\arraybackslash}p{0.2in}|>{\centering\arraybackslash}p{0.2in}|>{\centering\arraybackslash}p{0.5in}| }
 \hline
 System & Model & MRR & \multicolumn{2}{|c|}{Hits} & Time
 \\ 
& & & @1 & @10 & (min)\\
\hline
\hline
DGL-KE & Dot & .753 & .675 & .876 & 25.7m +-.17 \\
\hline
PBG & Dot & .751 & .672 & .873 & 23.6m +-.17\\ 
\hline
Marius & Dot & .750 & .672 & .872 & \textbf{12.5m} +-.01 \\
\hline
\end{tabular}
\caption{LiveJournal results after 25 epochs.}
\vspace{-.15in}
\label{tab:2}
\end{table}

\begin{table}[t]\scriptsize
\centering
  \begin{tabular}{ |c|c|>{\centering\arraybackslash}p{0.38in}|>{\centering\arraybackslash}p{0.2in}|>{\centering\arraybackslash}p{0.2in}|>{\centering\arraybackslash}p{0.34in}| }
 \hline
 System & Model & MRR & \multicolumn{2}{|c|}{Hits} & Time
 \\ 
& & & @1 & @10 & \\
\hline
\hline
PBG & Dot & .313 & .239 & .451 & 5h15m \\ 
\hline
 DGL-KE & Dot & .220 & .153 & .385 & 35h3m \\
\hline
Marius & Dot & .310 & .236 & .445 & \textbf{3h28m} \\
\hline
\end{tabular}
\caption{Twitter results after training for 10 epochs.}
\vspace{-.1in}
\label{tab:3}
\end{table}

\begin{table}[t]\scriptsize
\centering
  \begin{tabular}{ |c|c|>{\centering\arraybackslash}p{0.38in}|>{\centering\arraybackslash}p{0.2in}|>{\centering\arraybackslash}p{0.2in}|>{\centering\arraybackslash}p{0.34in}| }
 \hline
 System & Model & MRR & \multicolumn{2}{|c|}{Hits} & Time
 \\ 
& & & @1 & @10 & \\
\hline
\hline
PBG & ComplEx & .725 & .692 & .789 & 7h27m \\ 
\hline
Marius & ComplEx & .726 & .694 & .786 & \textbf{2h1m} \\
\hline
\end{tabular}
\caption{Freebase86m results with embedding size 100 after training for 10 epochs. }
\vspace{-.15in}
\label{tab:freebase86m}
\end{table}

\newparagraph{Twitter} We now move on to evaluating Marius on large-scale graphs for which embedding parameters do not fit in GPU memory. The Twitter follower network has approximately 1.4 billion edges and 41 million nodes. We train $100$-dimensional embeddings on each system using a dot product score function. We report results for one run for each system since we observed that training times and MRR are stable between runs. In total, there are 16 GB of embedding parameters with another 16 GB of optimizer state, since all systems use the Adagrad optimizer, as discussed above. To construct negatives for evaluation we use the approach from Zheng et al.~\cite{zheng2020dglke}, where 1,000 nodes are sampled uniformly from the graph, and 1,000 nodes are sampled by degree. 

Unlike the previous two datasets, each system uses a different methodology for training embeddings beyond GPU memory sizes. DGL-KE uses the approach described in Algorithm \ref{alg:SyncTraining}, storing parameters in CPU memory and processing batches synchronously while waiting for data movement. PBG does not utilize CPU memory and instead uses the partitioning approach with 16 partitions. Marius stores parameters in CPU memory, and utilizes its pipelined training architecture to overlap data movement with computation.

We compare the peak embedding quality learned by each of the systems after ten epochs of training in Table \ref{tab:3}. We find that Marius is able to train similar quality embeddings faster than the other systems, 10$\times$ faster than DGL-KE and 1.5$\times$ faster than PBG. DGL-KE's long training times can be attributed to data movement wait times inherent in synchronous processing. PBG on the other hand, only pays a data movement cost when swapping partitions. PBG achieves comparable runtimes because this dataset has a large amount of edges relative to the total number of parameters, meaning that computation times dominate partition swapping times. 

Turning our attention to the embedding quality, we find that Marius learns embedding of comparable quality to the next-best system: Marius yields an MRR of 0.310 versus 0.313 PBG. On the other hand, DGL-KE only achieves an MRR of 0.220. We attribute this gap in quality to to implementation differences between the systems (all use the same hyperparameters for training). 

\newparagraph{Freebase86m} We now evaluate Marius on a large-scale knowledge graph for which embedding parameters do not fit in CPU or GPU memory.  We train $100$-dimensional ComplEx embeddings for each system. In total, there are about 32 GB of embedding parameters with another 32 GB of Adagrad optimizer state. We do not evaluate DGL-KE on this dataset since it is unable to process this configuration on a single P3.2xLarge instance. For evaluation, we sample 1000 nodes uniformly and 1000 nodes based on degree as negative samples.

We compare the peak embedding quality of Marius and PBG where both systems are trained to 10 epochs in Table~\ref{tab:freebase86m}. Both systems use 16 partitions for training and in Marius we vary the number of partitions we hold in the CPU memory buffer. We find that Marius is able to train to peak embedding quality 3.7$\times$ faster when the buffer has a capacity of 8 partitions while reaching a similar accuracy. The runtime difference between the two systems can be attributed to the fewer number of partition swaps Marius performs and the ability to prefetch partitions.

\begin{figure}[t]
    \centering
    \includegraphics[width=0.48\textwidth]{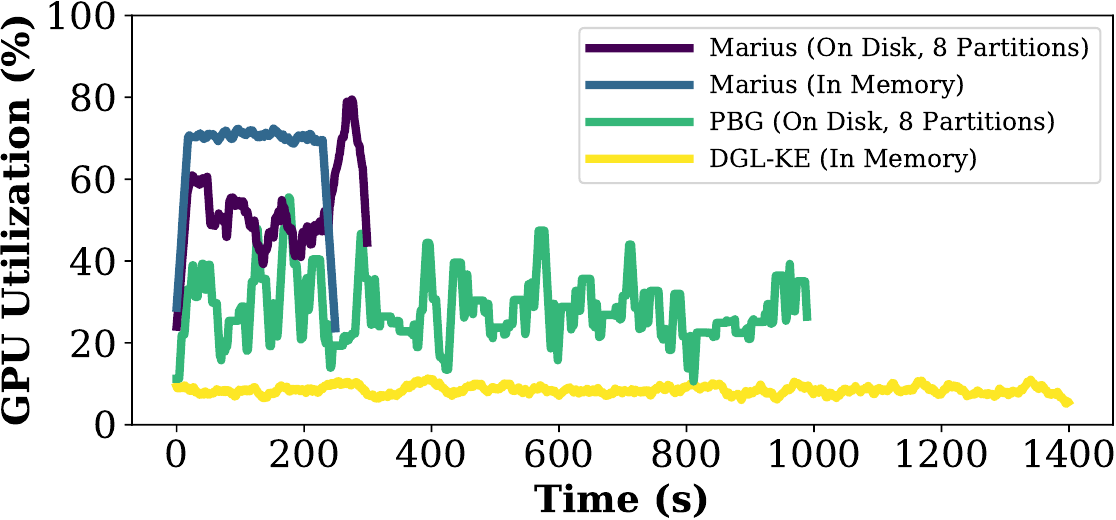}
    \vspace{-.2in}
    \caption{GPU utilization of Marius, DGL-KE and PBG during a single epoch of training $d=50$ embeddings on Freebase86m. Utilization is smoothed over 25 seconds. }
    \label{fig:utilization_marius}
    \vspace{-.15in}
\end{figure}

\newparagraph{Utilization} We include a comparison of GPU utilization during a single epoch of training $d=50$ embeddings on Freebase86m. Figure \ref{fig:utilization_marius} shows the utilization of two configurations of Marius compared to DGL-KE and PBG. One configuration of Marius stores embeddings in CPU memory while the other uses eight partitions on disk with four partitions buffered in CPU memory. We see that Marius is able to utilize the GPU 8$\times$ more than DGL-KE when training in memory and about 6$\times$ more when using the partition buffer. Compared to PBG, our partition buffer design leads to nearly 2$\times$ GPU utilization with fewer drops in utilization when waiting for partition swaps. While better than the baseline systems, Marius still doesn't achieve 100\% GPU utilization for this configuration. When profiling Marius with NVIDIA's nvprof, we found that all GPU operations were executed on the default CUDA stream, which is the default behavior of PyTorch. We plan to improve our implementation to leverage multiple CUDA streams thereby enabling GPU data transfer and compute to run in parallel, thereby improving GPU utilization. We have also found that the host CPU utilization could be a potential bottleneck (P3.2xLarge instance only has 8 vCPUs) and we plan to study techniques to mitigate CPU bottlenecks.

\newparagraph{Comparison vs. Distributed and Multi-GPU} We compare the training time and cost per epoch\footnote{All three systems converge in a similar number of epochs.} for Marius with the multi-GPU and distributed multi-CPU configurations of PBG and DGL-KE. PBG and DGL-KE support single machine multi-GPU training, and have a distributed multi-machine mode which is CPU-only. In the distributed configurations, the two systems partition parameters across the CPU memory of the machines and perform asynchronous training with CPU workers~\cite{pytorchbiggraph,zheng2020dglke}. Tables~\ref{tab:distributed50} and~\ref{tab:distributed100} show the configuration for each system and the corresponding epoch runtime and cost based on On-Demand AWS pricing. We observe that despite using a single GPU, Marius achieves comparable runtime with the multi-GPU configurations, while being more cost effective than all cases, ranging from 2.9$\times$ to 7.5$\times$ cheaper depending on the configuration. We also note that Marius can be extended to the multi-GPU setting; we discuss this in future work.
\begin{table}[t]\scriptsize
\centering
\begin{tabular}{ |c|c|c|c| } 
\hline
System & Deployment & Epoch Time (s) & Per Epoch Cost (\$) \\ 
\hline
\hline
Marius & 1-GPU & 288 & \textbf{.248} \\
\hline
DGL-KE & 2-GPUs & 761 & 1.29 \\
\hline
DGL-KE & 4-GPUs & 426 & 1.45 \\
\hline
DGL-KE & 8-GPUs & 220 & 1.50 \\
\hline
DGL-KE & Distributed & 1237 & 1.69 \\
\hline
PBG & 1-GPU & 1005 & .85 \\
\hline
PBG & 2-GPUs & 430 & .73 \\
\hline
PBG & 4-GPUs & 330 & 1.12 \\
\hline
PBG & 8-GPUs & 273 & 1.86 \\
\hline
PBG & Distributed & 1199 & 1.64 \\
\hline
\end{tabular}
\vspace{-.05in}
\caption{Cost comparisons with d=50 on Freebase86m.}
\vspace{-.05in}
\label{tab:distributed50}
\end{table}

\begin{table}[t]\scriptsize
\centering
\begin{tabular}{ |c|c|c|c| } 
\hline
System & Deployment & Epoch Time (s) & Per Epoch Cost (\$) \\ 
\hline
\hline
Marius & 1-GPU & 727 & \textbf{.61} \\
\hline
DGL-KE & 2-GPUs & 1068 & 1.81 \\
\hline
DGL-KE & 4-GPUs & 542 & 1.84 \\
\hline
DGL-KE & 8-GPUs & 277 & 1.88 \\
\hline
DGL-KE & Distributed & 1622 & 2.22 \\
\hline
PBG & 1-GPU & 3060 & 2.6 \\
\hline
PBG & 2-GPUs & 1400 & 2.38 \\
\hline
PBG & 4-GPUs & 515 & 1.75 \\
\hline
PBG & 8-GPUs & 419 & 2.84 \\
\hline
PBG & Distributed & 1474 & 2.02 \\
\hline
\end{tabular}
\vspace{-.05in}
\caption{Cost Comparisons with d=100 on Freebase86m.}
\vspace{-.15in}
\label{tab:distributed100}
\end{table}

\begin{figure}[t]
    \centering
    \includegraphics[width=0.48\textwidth]{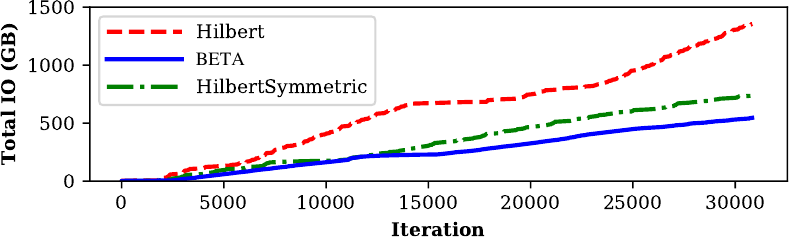}
    \vspace{-.2in}
    \caption{Total IO during a single epoch of training.}
    \vspace{-.15in}
    \label{fig:orderings_freebase86m.pdf}
\end{figure}

\subsection{Partition Orderings} \label{subsec:partition_ordering}

We now evaluate our buffer-aware \textit{BETA} ordering and compare it to two Hilbert curve based orderings. The first, \textit{Hilbert}, is the ordering generated directly from a Hilbert curve over the $n \times n$
matrix of edge buckets. The second, \textit{HilbertSymmetric}, modifies the previous curve by processing edge buckets $(i, j)$ and $(j, i)$ together, which reduces the overall number of swaps that need to be performed by about 2$\times$. All experiments use 32 partitions and a buffer capacity of 8 partitions.

\begin{figure*}[!tp]
    \centering
    \minipage[t]{0.33\textwidth}
    \includegraphics[width=0.83\linewidth]{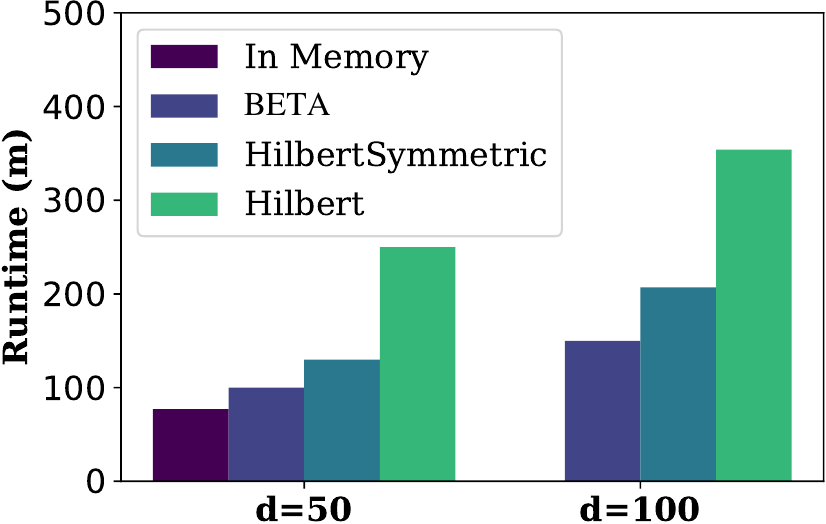}
    \vspace{-0.1in}
    \caption{10 epochs runtime per edge bucket ordering on Freebase86m.}
    \label{fig:fb86m_orderings_bar}
    \endminipage
    \hspace{0.05in}
    \minipage[t]{0.3\textwidth}
    \includegraphics[width=0.9\linewidth]{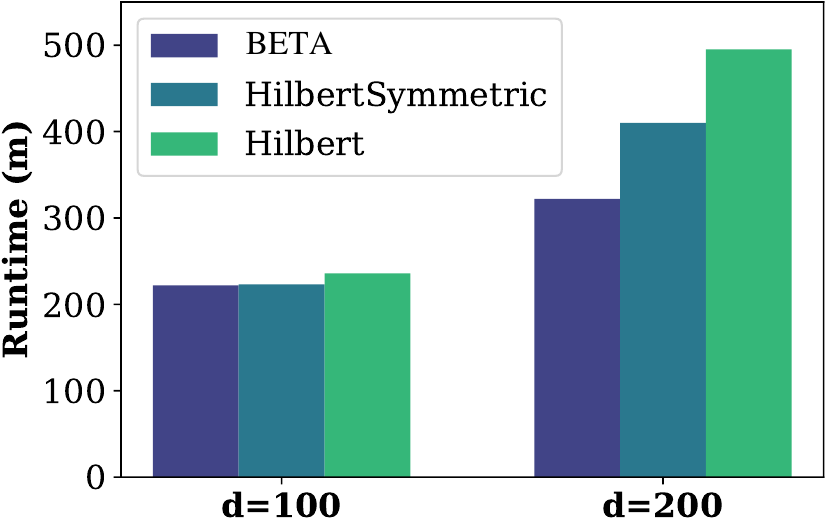}
    \vspace{-0.1in}
    \caption{10 epochs runtime per edge bucket ordering on Twitter.}
    \label{fig:twitter_orderings_bar}
    \endminipage
    \minipage[t]{0.35\textwidth}
    \includegraphics[width=\linewidth]{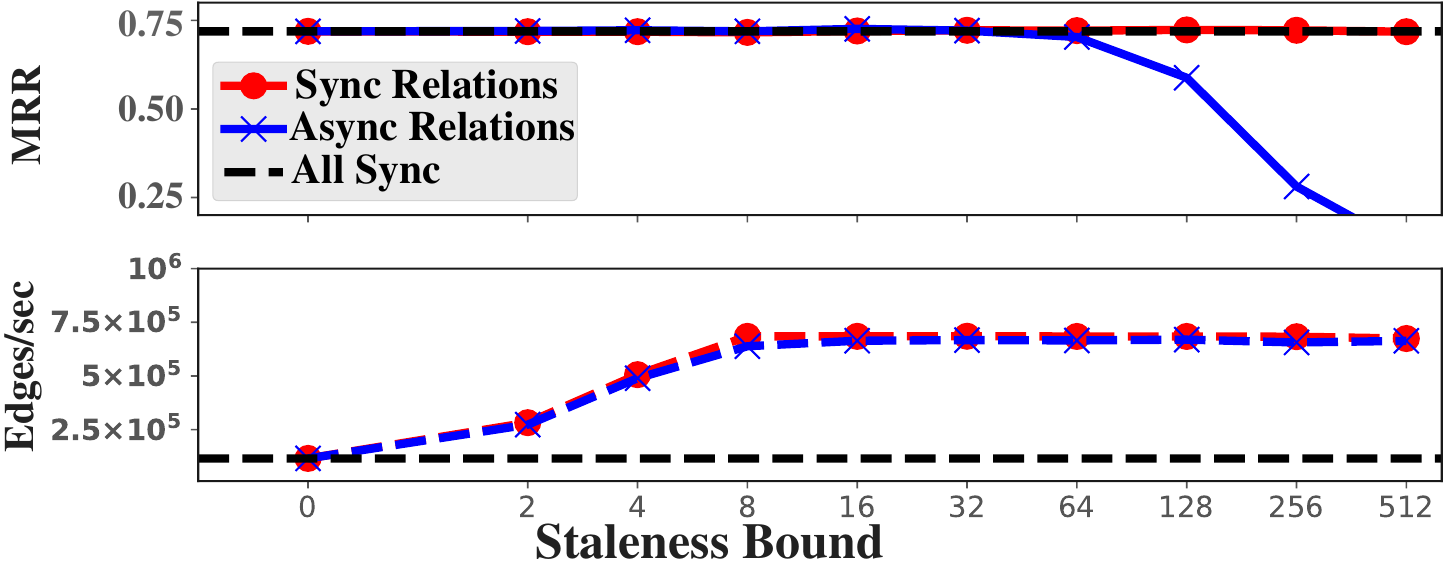}
    \vspace{-0.2in}
    \caption{Impact of staleness bound.}
    \label{fig:bounded_staleness}
    \endminipage\hfill
    \vspace{-0.1in}
\end{figure*}

We compare the orderings on Freebase86m with $d=50$ and $d=100$ sized embeddings, where the latter configuration exceeds CPU memory size. For $d=50$ we include an in-memory configuration which does not use partitioning as a baseline. Results are shown in Figure \ref{fig:fb86m_orderings_bar}. We find that the \textit{BETA} ordering reduces training time to nearly in-memory speeds, while only keeping 1/4 of the partitions in memory at any given time. The runtime of the three orderings is directly correlated with the amount of IO required to train a single epoch (Figure \ref{fig:orderings_freebase86m.pdf}). Since the \textit{Hilbert} and \textit{HilbertSymmetric} orderings require more IO, training stalls more often waiting for IO to complete. Results for $d=100$, also in Figure \ref{fig:fb86m_orderings_bar}, show that \textit{BETA} has the lowest training time, which is directly correlated with the amount of IO performed. Overall, the \textit{BETA} ordering is well suited for training large-scale graph embeddings through reducing IO. 

We also compare the orderings on Twitter with $d=100$ and $d=200$ sized embeddings. Results are shown in Figure \ref{fig:twitter_orderings_bar}. We find that the choice of ordering does not impact runtime for this configuration. Even though \textit{BETA} results in the smallest amount of total IO, the prefetching of partitions to the buffer always outpaces the speed of computation for the other orderings. We see this in Twitter and not Freebase86m, because Twitter has nearly 10$\times$ the density of Freebase86m, i.e., more computation needs to be performed per partition. When we increase the embedding dimension to $d=200$ (Figure \ref{fig:twitter_orderings_bar}) we see a difference in running time. By increasing the embedding dimension by 2$\times$ we increase the total amount of IO by 2$\times$, and now the prefetching of partitions is outpaced by the computation. 

Overall, we see that certain configurations are \textit{data bound} and others are \textit{compute bound}. For data bound configurations like $d=50$ and $d=100$ on Freebase86m and $d=200$ on Twitter, the choice of ordering will impact overall training time, with \textit{BETA} performing best. But for \textit{compute bound} workloads such as $d=100$ on Twitter, the choice of ordering makes little difference since the prefetching always outpaces computation. 

\subsection{Large Embeddings} \label{subsec:large_embeddings}

We evaluate the ability of Marius to scale training beyond CPU memory sizes in this section. We vary the embedding dimension from a small dimension of $d=20$, for which training fits in GPU memory, to a large embedding dimension $d=800$, which is well beyond the memory capacity of a single P3.2xLarge instance. The results are shown in Table~\ref{tab:big_embeddings}. We find that the embedding quality increases with increased embedding dimension. We also see that as the embedding size increases, the training time increases quadratically. We see this because the number of swaps and total IO scales quadratically with the number of partitions, if the buffer capacity is held fixed. And because training is bottlenecked by IO for large embedding sizes, we see quadratic runtime increases. It should be noted that with a faster disk we would observe improved runtimes, with a 2$\times$ faster disk leading to 2$\times$ faster training for large embeddings. With NVMe-based SSDs becoming more common, the design of Marius will best be able to leverage future fast sequential storage mediums and scale training to embedding sizes beyond what we show here.

\begin{table}[t]\scriptsize
    \centering
    \begin{tabular}{ |c|c|c|c|c| } 
        \hline
        d & Size & Partitions & MRR & Runtime (Epoch)   \\ 
        \hline
        \hline
        20 & 13.6 GB & - & .698 & 4m \\ 
        \hline
        50 & 34.4 GB & - & .722 & 4.8m \\ 
        \hline
        100 & 68.8 GB & 32 & .726 & 12.1m \\ 
        \hline
        400 & 275.2 GB & 32 & .731 & 92.4m \\ 
        \hline
        800 & 550.4 GB & 64 & .731 & 396m \\ 
        \hline
    \end{tabular}
    \caption{Freebase86m. $d=400$ and $d=800$ trained to 5 epochs, other cases are trained to 10. }
    \vspace{-.15in}
    \label{tab:big_embeddings}
\end{table}

\subsection{Microbenchmarks}
\label{sec:exp_microbenchmarks}

\newparagraph{Bounded Staleness} 
We now show how our pipelined training architecture with bounded staleness affects the embedding quality and throughput of training. We train Marius on Freebase86m with $d=50$, and vary the number of batches allowed into the pipeline at any given time. We evaluate how the performance and MRR vary as we vary the staleness bound. We compare three cases, synchronous updates to all parameters, synchronous updates to only the relation embeddings, and asynchronous updates to all parameters.\footnote{For asynchronous updates to the relation embeddings, we pipe them to the GPU from CPU memory as with the node embeddings.} Results are shown in Figure~\ref{fig:bounded_staleness}. We see that increasing the staleness bound when asynchronously updating the relation embeddings results in severe degradation of embedding quality. For synchronous updates of the relation embeddings and asynchronous updates to the node embeddings, we see that MRR does not degrade significantly with increasing staleness bound. This suggests that relation embeddings are sensitive to staleness, which might be due to dense updates. These results additionally show that node embeddings are not sensitive to asynchronous updates, which may be due to sparse updates. We also find that increasing the bound improves the throughput of the system by about a factor of 5 over synchronous training but that the benefits diminish beyond a staleness bound of 8.

\begin{figure}[t]
    \vspace{-0.075in}
    \centering
    \includegraphics[width=0.48\textwidth]{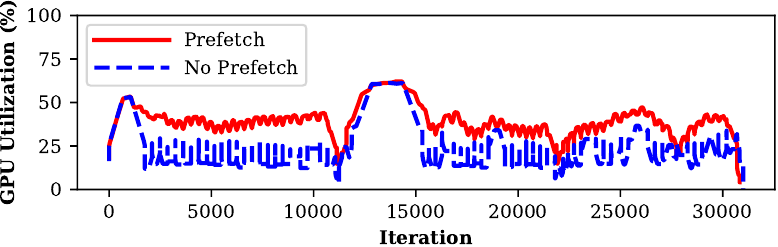}
    \vspace{-0.2in}
    \caption{Effect of prefetching with Freebase86m.}
    \vspace{-.15in}
    \label{fig:prefetch}
\end{figure}

\newparagraph{Prefetching Effects}
We evaluate the effect of prefetching partitions to the buffer on GPU utilization. We train Marius on Freebase86m with $d=100$, 32 partitions, and a buffer capacity of eight. We show the average utilization of the GPU during each iteration of a single epoch of training in Figure~\ref{fig:prefetch}. We can see that prefetching results in a higher sustained utilization of the GPU since less time is spent waiting for partition swaps. Interestingly, both configurations see a utilization bump starting at about iteration 12,000. This is because the \textit{BETA} ordering does not require any swaps during this period. Overall, prefetching is able to mitigate wait times for partition swaps improving utilization and training times.

\section{Discussion}
We next discuss some lessons learned from Marius deployments in the cloud and discuss how the \textit{BETA} ordering aims to optimize a workload that is fundamentally different than those considered by prior large-scale graph processing paradigms.
\subsection{Deployment Considerations} 
Given the diverse set of cloud computing instances offered by vendors, there are a wide variety of possible hardware deployments with associated costs and benefits. It is challenging to determine what the best deployment option is, especially with performance also being impacted by the choice of model and dataset. Here we list some considerations when deploying Marius.

\newparagraph{Properties of the Input Graph} Training time and storage overhead are largely driven by the size of the input graph. More edges lead to more computation, and more nodes and edge-types results in a larger storage footprint. The density of the graph impacts the bottleneck of the system when using the partition buffer. A graph with high density will be compute bound, as more computation will have to be performed on each node partition, as we see in Figure~\ref{fig:twitter_orderings_bar}. For sparse graphs the training will be data bound, as we see in Figure~\ref{fig:fb86m_orderings_bar}. For data bound settings, utilizing a storage device with high throughput can improve training times, while for compute bound workloads, more GPUs and parallelism can help. 
 
\newparagraph{Model Complexity} Some models such as \textit{DistMult}, \textit{ComplEx}, and \textit{Dot} are computationally simple, only requiring dot products and element-wise multiplication, while others such as CapsE~\cite{nguyen2019capsule} utilize convolutions and a capsule neural network. Training simple models requires less compute to perform the forward and backwards pass and therefore is more likely to be data bound. The opposite is true for complex models. 

\newparagraph{Configuration and Tuning} A major challenge with training graph embedding models is the number of hyperparameters which impact embedding quality, training throughput, and convergence rates. In terms of batch size, we observe that large batches ($\approx$10000) can improve training throughput with no impact on model accuracy for large graphs, but throughput benefits diminish after a certain batch size. Throughput can also be increased by using a larger number of partitions (Figure~\ref{fig:greedy-io}), but this affects embedding quality. IO can also be reduced by increasing the capacity of the buffer, which quadratically reduces the number of swaps; thus it is best to size the buffer to the maximum number of partitions that will fit in CPU memory. Finally, as seen in Figure \ref{fig:bounded_staleness}, increasing the staleness bound improves training throughput but can negatively impact embedding quality. Overall, the effect of these parameters are graph dependent, and efficiently tuning hyperparameters for a given graph is an interesting direction for future work.

\subsection{Out-of-core Graph Processing} 
The graph embedding workload requires iterating over edges and computing on data associated with both end-points (i.e., the embeddings of source, destination). The \textit{BETA} ordering is designed to minimize IO when accessing node embedding vectors associated with edges that are being processed. Classic graph processing systems and methods such as GraphChi’s Parallel Sliding Window (PSW) \cite{10.5555/2387880.2387884} are tailored for workloads that iterate over vertices and process data associated with the incoming edges of each node. Applying such schemes (e.g., PSW) to graph embeddings would require performing redundant IO (scaling quadratically with partitions) to access embeddings for both incoming/outgoing vertices. Furthermore, for classic graph processing algorithms such as PageRank, the storage overhead of node data is only a single float or a low dimensional vector. Based on this, traditional graph processing systems make the assumption that storing and accessing node data is inexpensive and fits in memory. In contrast, graph embeddings are high dimensional vectors making storing and accessing node data costly, and hence the workload requires new graph traversal algorithms to minimize IO.

\section{Conclusion}
We introduced Marius, a new framework for computing large-scale graph embedding models on a single machine. We demonstrated that the key to scalable training of graph embeddings is optimized data movement. To optimize data movement and maximize GPU utilization, we proposed a pipelined architecture that leverages partition caching and the {\em BETA ordering}, a novel buffer-aware data ordering scheme. We showed using standard benchmarks that Marius achieves the same accuracy but is up to an order-of magnitude faster than existing systems. We also showed that Marius can scale to graph instances with more than a billion edges and up to 550 GB of model parameters on a single AWS P3.2xLarge instance. In the future, we plan to explore how the ideas behind Marius' design and our new data ordering can be applied to  distributed setting and help speed up training of graph neural networks.

\paragraph{Acknowledgements}
This work was supported by NSF under grant 1815538 and DARPA under grant ASKE HR00111990013. The U.S. Government is authorized to reproduce and distribute reprints for Governmental purposes notwithstanding any copyright notation thereon. Any opinions, findings, and conclusions or recommendations expressed in this material are those of the authors and do not necessarily reflect the views, policies, or endorsements, either expressed or implied, of DARPA or the U.S. Government. This work is also supported by the National Science Foundation grant CNS-1838733, a Facebook faculty research award and by the Office of the Vice Chancellor for Research and Graduate Education at UW-Madison with funding from the Wisconsin Alumni Research Foundation.

\newpage

\bibliographystyle{plain}
\bibliography{ref}

\newpage

\appendix
\section{Artifact Appendix}

\subsection*{Abstract}

The artifact includes the Marius source code, configuration and scripts for all experiments, including the baselines. Details on how to use the artifact can be found in the README file in our Github repository.

\subsection*{Scope}

The artifact can be used to validate and reproduce the results for all experiments. The source code and experiment configuration can be viewed to obtain any implementation details that were not mentioned in the paper for brevity. We do not include the source code to PyTorch Big-Graph and DGL-KE, the versions used in this work can be found at:
\begin{small}
\url{https://github.com/facebookresearch/PyTorch-BigGraph/tree/4571deee78d0fff974a81312c0c3231d7dc96a69}
\end{small} 
and 
\begin{small}
\url{https://github.com/awslabs/dgl-ke/releases/tag/0.1.1} 
\end{small} 

\subsection*{Contents}

\newparagraph{Marius:} The source code for Marius is mostly written in C++ with bindings to support a Python API. Located in \nolinkurl{/src}. The version of Marius in the artifact (\texttt{osdi2021} branch) corresponds to the version used to produce results in this paper. We are also actively making improvements to Marius and the latest version of the system can be found in the main branch of the repository. The main branch of the repository also contains documentation on how to use Marius directly. 

\newparagraph{Experiment runner:} A collection of Python scripts that can be used to run the experiments used in the paper. Located in \nolinkurl{/osdi2021}. Experiments can be run with \nolinkurl{python3}  \nolinkurl{osdi2021/run_experiment.py}  \nolinkurl{--experiment} \nolinkurl{<EXPERIMENT>}. A full list of experiments can be found by passing the \nolinkurl{--help} flag. Once an experiment has run to completion, results are output to the terminal and detailed results, metrics and figures can be found in the experiment directory.

\newparagraph{Experiment configuration:} Configuration files for all experiments and baselines are stored in their corresponding experiment directory. The experiment directories are named based on their corresponding experiment in this paper. For example, the experiment configuration files and results after running the experiments for Section \ref{sec:system_comparisons} can be found in \nolinkurl{/osdi2021/system_comparisons/<DATASET>/<SYSTEM/}

\newparagraph{Datasets and preprocessing code:} Code to download and preprocess the datasets used in this paper are included in the artifact. For Marius they can be found in \nolinkurl{/src/python/tools}. For DGL-KE and PBG, they can be found in \nolinkurl{/osdi2021/dglke_preprocessing} and \nolinkurl{/osdi2021/pbg_preprocessing}. We used four datasets in our evaluation (Table \ref{tab:datasets}). FB15k and Freebase86m are a subset of the Freebase knowledge graph \cite{freebase}, where each edge is encoded as a triple. Each triple encodes general factual information. An example triple is of the form: \nolinkurl{(Giannis} \nolinkurl{Antetokounmpo,} \nolinkurl{Plays,} \nolinkurl{Basketball)}. LiveJournal \cite{livejournal} is a social network dataset where nodes represent users and edges denote friendships between them. Similarly, the Twitter \cite{twitter} dataset contains a follower network between users. 

\newparagraph{Buffer simulator:} The buffer simulator was used to develop and test edge bucket orderings. It computes the number of swaps for any edge bucket ordering for any number of partitions and any buffer size.

\subsection*{Hosting}

\newparagraph{Artifact:} \\ \url{https://github.com/marius-team/marius/tree/osdi2021}

 \newparagraph{Latest version:} \\ \url{https://github.com/marius-team/marius}

\subsection*{Requirements}

Detailed software requirements and dependencies are listed in the artifact README. The artifact must be run on a machine with an NVIDIA GPU. The target deployment for this artifact is the P3.2xLarge instance from AWS. There are a few experiments which cannot run on this instance due to memory limitations. We detail these in the README. 

\end{document}